\pdfoutput=1

\documentclass[11pt]{article}

\usepackage[preprint]{acl}
\usepackage{times}
\usepackage{latexsym}

\usepackage[T1]{fontenc}

\usepackage[utf8]{inputenc}

\usepackage{microtype}

\usepackage{inconsolata}

\usepackage{graphicx}

\title{SafeSynthDP: Leveraging Large Language Models for Privacy-Preserving Synthetic Data Generation Using Differential Privacy}

\author{Md Mahadi Hasan Nahid \\
  University of Alberta\\
  \texttt{mnahid@ualberta.ca} \\\And
  Sadid Bin Hasan \\
  University of Alberta\\
  \texttt{sadidbin@ualberta.ca}\\}

\begin{document}
\maketitle

\begin{abstract}

Machine learning (ML) models frequently rely on training data that may include sensitive or personal information, raising substantial privacy concerns. Legislative frameworks such as the General Data Protection Regulation (GDPR) and the California Consumer Privacy Act (CCPA) have necessitated the development of strategies that preserve privacy while maintaining the utility of data. In this paper, we investigate the capability of Large Language Models (LLMs) to generate synthetic datasets integrated with Differential Privacy (DP) mechanisms, thereby enabling data-driven research and model training without direct exposure of sensitive information. Our approach incorporates DP-based noise injection methods, including Laplace and Gaussian distributions, into the data generation process. We then evaluate the utility of these DP-enhanced synthetic datasets by comparing the performance of ML models trained on them against models trained on the original data. To substantiate privacy guarantees, we assess the resilience of the generated synthetic data to membership inference attacks and related threats. The experimental results demonstrate that integrating DP within LLM-driven synthetic data generation offers a viable balance between privacy protection and data utility. This study provides a foundational methodology and insight into the privacy-preserving capabilities of LLMs, paving the way for compliant and effective ML research and applications. 

\end{abstract}

\section{Introduction} 

The increasing reliance on machine learning (ML) models for data-driven decision-making across sectors—ranging from healthcare and finance to social media content moderation—has amplified concerns surrounding data privacy. Models often require training on sensitive, personal information, heightening the risk of inadvertently exposing confidential details. Legislative frameworks, such as the General Data Protection Regulation (GDPR) \cite{marjanov2023data, nouwens2020dark} and the California Consumer Privacy Act (CCPA) \cite{samarin2023lessons}, strictly regulate the handling of personal data, making it imperative for researchers and practitioners to explore strategies that protect individual privacy without compromising model performance. These stringent standards motivate our exploration of privacy-preserving synthetic data generation, where the goal is to produce artificial datasets that mirror the statistical characteristics of real data but do not reveal identifiable information.

Recent advances in Large Language Models (LLMs) have opened novel avenues for synthetic data generation. LLMs excel at capturing complex distributions, patterns, and linguistic structures from diverse corpora. Beyond traditional ML tasks, LLMs are increasingly used for processes like \emph{fine-tuning}—adjusting pretrained models with domain-specific data—and \emph{in-context learning} (ICL)—guiding model behavior by providing a few annotated examples as prompts at inference time \cite{brown2020languagemodelsfewshotlearners, wei2022chain, NEURIPS2020_1457c0d6}. However, incorporating sensitive data during either fine-tuning or ICL introduces privacy risks, as LLMs may memorize and subsequently disclose private information \cite{tang2024privacypreserving, hong2024dpopt}. To mitigate this risk, we propose integrating Differential Privacy (DP) \cite{dwork2014algorithmic} directly into the synthetic data creation pipeline. By adding mathematically calibrated noise to the generation process, we ensure that no single individual’s information unduly influences the model’s output, significantly reducing the probability of re-identifying or inferring private details.

A practical example of the necessity for such techniques can be found in the healthcare domain. Medical researchers often train ML models on electronic health records (EHRs) to predict patient outcomes or recommend treatments. Suppose an ML model, trained directly on real patient data, learns to identify rare conditions. If prompted (either directly or inadvertently during inference), it could reveal sensitive patient attributes, thereby violating privacy regulations and ethical standards. Instead, consider generating a DP-enhanced synthetic dataset that statistically resembles the EHRs but omits any identifying details. This synthetic dataset can still support tasks like text classification (e.g., categorizing clinical notes or predicting risk levels) and improve models through ICL without endangering patient privacy \cite{keshta2021security}. 

Synthetic datasets play a crucial role in this context, proving invaluable for the development of domain-specific language models, such as those designed for educational purposes like a Language Model for Kids \cite{nayeem-rafiei-2024-kidlm, liu2024best}. A vast majority of data are stored in textual formats, either structured or semi-structured, like tables \cite{brown2020languagemodelsfewshotlearners, nahid2024improving}. Large Language Models also exhibit significant capabilities in textual comprehension, allowing them to process and reason over various text-based data, including structured tabular data in various downstream tasks such as processing, prediction, and question-answering \cite{nahid-rafiei-2024-normtab, nahid-rafiei-2024-tabsqlify}.

Differential Privacy (DP) provides a principled framework for protecting individual-level information in datasets. By adding carefully calibrated noise to computations, DP ensures that the inclusion or exclusion of a single data point does not significantly affect the aggregate output of a process, thereby limiting the risk of inference attacks \cite{shokri2017membership}. While DP has been extensively studied in the context of traditional data analysis and model training, its direct integration into the synthetic data generation process, guided by LLMs, remains a relatively unexplored avenue. This integration enables the production of synthetic datasets that resemble the statistical properties of real data while obscuring individual-level details, ensuring compliance with privacy regulations and ethical mandates.

LLMs, which have demonstrated remarkable capabilities in capturing complex distributions and linguistic structures, present a unique opportunity for generating DP-enhanced synthetic data. By employing LLMs, it is possible to produce realistic, domain-relevant data surrogates without repeatedly exposing the original sensitive datasets. When combined with DP mechanisms, this approach aims to safeguard privacy from the inception of data generation. The resulting synthetic datasets can be used for downstream ML tasks, such as classification, without sacrificing performance or privacy integrity.
This research is structured around three key questions that guide our exploration of integrating DP within LLM-driven synthetic data generation:

\begin{enumerate} \item \textbf{Preservation of Model Utility:} To what extent can DP-enhanced synthetic datasets produced by LLMs preserve the predictive accuracy and robustness of ML models relative to models trained on real data? \item \textbf{Balancing Privacy and Utility:} How do different DP noise parameters and distributions affect the trade-off between privacy guarantees and data utility, and under which conditions can acceptable performance be maintained? \item \textbf{Generality Across ML Architectures:} Can DP-enhanced synthetic datasets support a wide range of ML, Deep Learning (DL), and advanced LLM-based methods, thereby generalizing the applicability of this privacy-preserving data generation paradigm? \end{enumerate}
This study makes several noteworthy contributions to the field of privacy-preserving ML:

\begin{itemize} \item \textbf{Integration of DP into LLM-Based Synthetic Data Generation:} We present a novel framework,  \textbf{\textsc{SafeSynthDP}} that merges DP principles with LLM-driven data generation, enabling the creation of synthetic datasets that emulate sensitive information distributions without revealing individual-level details. 

\item \textbf{Quantitative Evaluation of Privacy-Utility Trade-offs:} We provide a rigorous empirical assessment of various DP parameters, including privacy budgets and noise distributions, to establish guidelines for achieving an optimal balance between privacy and data utility. 

\item \textbf{Broad Architectural Validation:} Our methodology is validated across multiple ML architectures—from traditional algorithms to sophisticated DL models (e.g., GRU, LSTM) and state-of-the-art LLMs (e.g., gpt-4o-mini \footnote{\href{https://platform.openai.com/docs/models/gpt-4o-mini}{https://platform.openai.com/docs/models/gpt-4o-mini}}, gemini-1.5-flash \footnote{\href{https://deepmind.google/technologies/gemini/flash/}{https://deepmind.google/technologies/gemini/flash/}}) demonstrating the generality and practical relevance of our approach. 
\end{itemize}

In the following sections, we first review the related works on prompting, in-context learning, DP, and synthetic data generation \footnote{For background see Appendix \ref{appendix: background}.}. We then detail our methodology, experimental setup, results, and analysis, before concluding with a discussion of the implications and potential future directions of this research.

\section{Related Work}
Developing effective methods for privacy-preserving data release has long been a core challenge in machine learning research, particularly as organizations increasingly leverage real-world datasets containing sensitive information. Over time, a variety of approaches have emerged to generate synthetic datasets with Differential Privacy (DP), striving to protect individual records while retaining the statistical properties necessary for downstream analysis and modeling. This section surveys key contributions in DP-based synthetic data generation and highlights how existing methods motivate our approach, which uniquely integrates Large Language Models (LLMs) to produce utility-preserving, privacy-compliant datasets.

\paragraph{Balancing Privacy and Utility in Synthetic Data Generation.}  
Early investigations into synthetic data as a privacy solution grappled with the fundamental tension between utility and protection. For instance, Stadler et al.~\cite{stadler2022synthetic} demonstrated that while synthetic data often outperforms naive anonymization methods, achieving a stable trade-off between preserving meaningful information and safeguarding individual privacy remains difficult. Building upon this premise, researchers have sought more principled solutions that employ DP mechanisms directly within the data generation process. Zhang et al.~\cite{zhang2021privsyn}, in their work on PrivSyn, embedded DP into the synthesis of high-dimensional tabular data, ensuring that nuanced attribute correlations are maintained without disclosing sensitive records. Similarly, Zhang et al.~\cite{zhang2017privbayes} proposed PrivBayes, leveraging Bayesian networks to preserve essential statistical relationships and achieve strong DP guarantees, thus offering an early blueprint for balancing complexity, dimension, and privacy rigor.

\paragraph{Refining Techniques for Enhanced Utility and Applicability.}  
As research advanced, emphasis shifted toward refining synthetic data methods that bolster utility and scalability. Arnold and Neunhoeffer~\cite{arnold2020differentially} explored ensemble strategies, such as QUAIL, to combine multiple DP-based generators and enhance dataset quality for machine learning tasks like classification. Concurrently, Long et al.~\cite{long2021g_pate} introduced G-PATE, integrating Private Aggregation of Teacher Ensembles with GANs \cite{goodfellow2014generative}, showing that when carefully tuned, synthetic data can yield performance close to that of real datasets. These works underscore that improved privacy-utility calibrations are possible through advanced algorithmic formulations, hinting at the potential of more flexible generation techniques that adapt to different data distributions and task requirements.

\paragraph{Addressing Complexity and Theoretical Boundaries.}  
While certain methods performed well in controlled settings, researchers recognized that practical constraints and intricate data distributions still pose hurdles. For instance, Cai et al.~\cite{10.14778/3476249.3476272} introduced PrivMRF, using Markov Random Fields to model correlations in structured data. Despite strong results for counting and classification, PrivMRF struggled with highly complex dependencies. Hardt et al.~\cite{NIPS2012_208e43f0} proposed early, more generic DP data release algorithms that were simple yet faced scalability limits. Rosenblatt et al.~\cite{rosenblatt2020differentiallyprivatesyntheticdata} and Torkzadehmahani et al.~\cite{torkzadehmahani2019dp} evaluated and developed synthetic data generation techniques like DP-CGAN, calling for enhancements in both model stability and output fidelity. Vietri et al.~\cite{pmlr-v119-vietri20b} explored oracle-efficient algorithms that achieved accuracy but depended heavily on advanced optimization tools. Collectively, these investigations illustrated that while DP-driven synthetic data methods could yield substantial gains, they often encountered bottlenecks in complexity, stability, or computational overhead.

\paragraph{Evolving from Model-Centric Approaches to Flexible Frameworks.}  
Most early efforts integrated DP through model-centric approaches, focusing on gradient perturbation (e.g., DP-SGD)~\cite{dwork2014algorithmic,10.1145/2976749.2978318}, or applied noise during post-processing steps. While such methods reduced privacy risks, they frequently impaired data utility and model performance~\cite{10.14778/3611479.3611517}. Moreover, they typically required extensive training resources and struggled to adapt to rapidly changing tasks. These challenges highlight the need for approaches that minimize computational strain while maintaining rigorous privacy standards. The literature to date suggests that versatile and adaptive synthetic data methods—ones that can incorporate domain-specific constraints, handle complex patterns, and maintain high fidelity—are essential for real-world deployments.

\paragraph{Towards LLM-Driven Synthetic Data Generation.} ~Despite the progress in DP-based synthetic data methods, relatively little attention has been devoted to leveraging Large Language Models (LLMs) for dataset generation. LLMs, such as GPT-4 \cite{achiam2023gpt}, have shown remarkable capabilities in capturing contextual and linguistic patterns, but their integration into DP workflows remains nascent. Although some studies have examined DP during LLM inference \cite{NEURIPS2020_1457c0d6}, and prompt engineering techniques have been proposed to safeguard sensitive inputs \cite{tang2024privacypreserving, hong2024dpopt, chua2024mind}, the literature has yet to fully explore LLMs as generative engines for DP-compliant synthetic datasets. Addressing this gap is crucial, as LLMs can produce highly contextualized, linguistically coherent data that might better reflect real-world distributions without disclosing private attributes.

\paragraph{Motivation for Our Approach.}  
Our work draws direct inspiration from these bodies of research. From early efforts, we learn the importance of balancing privacy and utility when producing synthetic datasets. From more recent advances, we adopt the idea that careful integration of DP can yield robust utility even in challenging settings. By leveraging LLMs, we aim to push beyond limitations encountered by conventional generative models, such as instability or excessive computational overhead, and generate synthetic data that preserves critical patterns needed for machine learning tasks—particularly classification—while ensuring strong privacy protections. In essence, our approach strives to unite the analytical rigor of DP frameworks with the adaptability and expressiveness of LLMs, taking a decisive step toward producing practical, privacy-preserving synthetic datasets that align with stringent regulatory and ethical standards.

\section{Methodology}

\begin{figure*}[ht]
    \centering
    \includegraphics[width=0.8\linewidth]{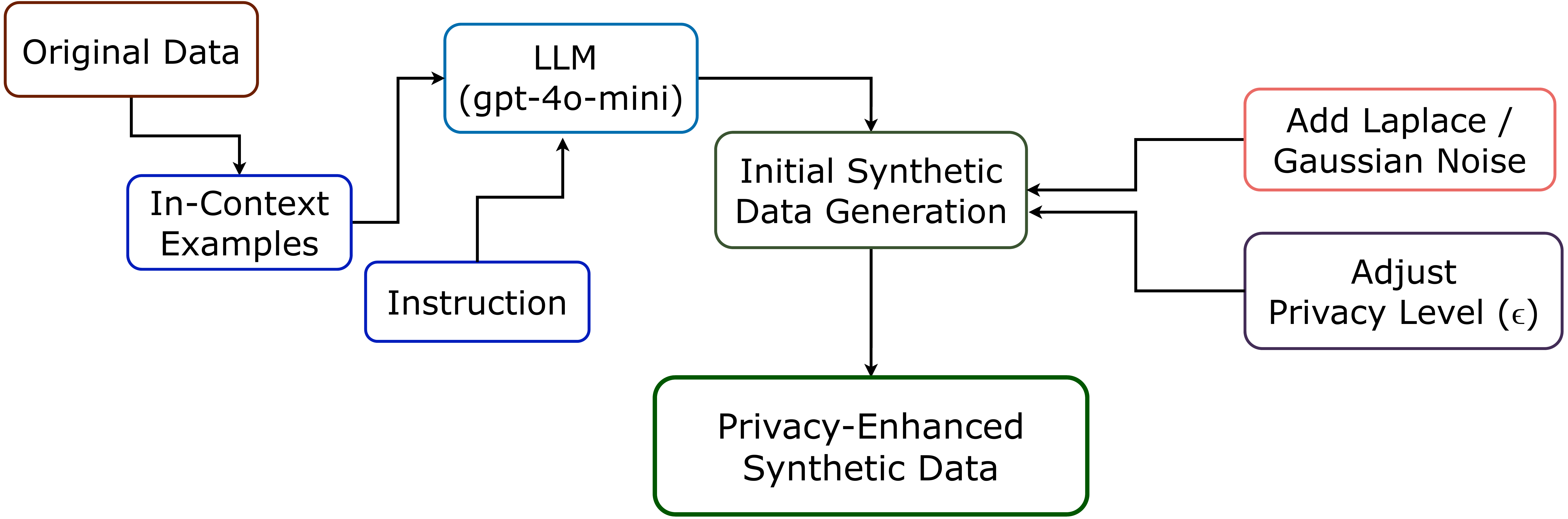}
    \caption{Workflow for Generating Privacy-Preserving Synthetic Data (\textbf{SafeSynthDP}). This diagram illustrates the process from selecting in-context examples from the original dataset, through the generation of initial synthetic data using the gpt-4o-mini LLM, to enhancing privacy through the addition of Laplace or Gaussian noise and adjusting the privacy level via the  $\epsilon$ parameter, culminating in the evaluation of the privacy-enhanced synthetic data.}
    \label{fig:process}
\end{figure*}

Our methodology for generating differentially private synthetic data involves utilizing the advanced capabilities of Large Language Models (LLMs) to create datasets that closely mimic the statistical and semantic characteristics of an original, private dataset. Unlike traditional methods which might involve extensive training or fine-tuning, our approach is training-free, which significantly reduces computational demands and avoids the direct exposure of sensitive information from the source dataset. We detail below our process for generating and evaluating these synthetic datasets, focusing on integrating Differential Privacy (DP) mechanisms to bolster privacy protection.

To ensure that the synthetic data generated by the LLM retains both the structural integrity and thematic essence of the original data, we use a demonstration-based prompting strategy. We select a few examples from the dataset to serve as in-context demonstrations within the prompt. These examples guide the LLM to replicate the style, topics, and language patterns of the original data without directly reproducing it. By providing these exemplars, we instruct the model to generate new content that captures the statistical distribution and contextual relevance of the original dataset. For this task, we employ two recent LLMs from different model families: gpt-4o-mini \cite{hurst2024gpt} from the GPT series and gemini-1.5-flash \cite{team2024gemini} from Google's Gemini suite. This allows us to compare the effectiveness of synthetic data generation across different model architectures.

In our approach, we integrate Differential Privacy (DP) mechanisms directly into the data generation process for enhanced privacy. We use Laplace and Gaussian noise mechanisms to inject controlled noise into the synthetic data based on the distribution of the words. Laplace noise is applied to each data point or feature, which helps maintain the utility of the data while providing privacy. The amount of noise in Laplace mechanism or Gaussian mechanism for differential privacy is calculated based on the Laplace and Gaussian distribution equation. 

The level of noise is inversely proportional to the privacy budget $\epsilon$,  where a lower $\epsilon$ indicates stronger privacy at the cost of data utility. On the other hand, Gaussian noise, known for its smoother distribution, is particularly useful for preserving the integrity of continuous data features. The noise parameters are adjusted based on the sensitivity of the data and the desired privacy level. These mechanisms are critical as they ensure that even if an adversary tries to infer membership or reconstruct original data points, the added noise significantly obscures individual entries, protecting against both membership inference attacks and attribute disclosure while maintaining data utility for machine learning tasks.

For the evaluation process, we conduct comparative experiments where we train both simple ML models, like Multinomial Naive Bayes \cite{10.1007/978-3-540-30549-1_43} and Support Vector Machines \cite{708428}, and complex deep learning models, such as Gated Recurrent Units \cite{chung2014empirical} and Long Short-Term Memory networks \cite{10.1162/neco.1997.9.8.1735}, on both the original and synthetic datasets. We then evaluate their performance on a held-out test set from the original data. Additionally, we assess how well the synthetic data performs in in-context learning scenarios with the LLMs used for generation, providing insights into its ability to mimic real data for advanced learning tasks.

\subsection{Synthetic Data Generation Process}

Our approach to generating synthetic data leverages the capabilities of the gpt-4o-mini model, which we use to produce datasets that emulate the structural and thematic qualities of the original data while ensuring privacy through the application of the Laplace mechanism. The process is summarized in \ref{fig:process}. The process is articulated in the following key steps:\\

\textbf{Generating Synthetic Data with LLM:} 
We utilize in-context learning by providing gpt-4o-mini with a curated set of demonstration prompts or examples drawn from the original dataset. For instance, when the goal is to generate synthetic news headlines, we supply the model with exemplar headlines that capture various news themes, formats, and linguistic styles. This guidance directs gpt-4o-mini to create analogous content, effectively crafting a dataset that statistically and contextually resembles the original without directly replicating any sensitive information. \\ 

\textbf{Privacy Enhancement through DP Mechanisms:} 
To further enhance privacy, we integrate Laplace and/or Gaussian mechanisms during the data generation phase with LLM. We manipulate token frequencies in the generated text by adding carefully calibrated noise based on the Laplace or Gaussian distribution. For the Laplace mechanism, we adjust the counts of common words or phrases, altering their likelihood in the synthetic text. Similarly, Gaussian noise is applied with a distribution that helps smooth the impact on the text's features. These noise additions are pivotal in masking any identifiable patterns or information that could be exploited in privacy attacks, thereby strengthening the privacy safeguards of the dataset. \\ 

\textbf{Adjusting Hyperparameters for Privacy Level ($\epsilon$):} 
The privacy level is governed by the hyperparameter $\epsilon$ (epsilon). A smaller $\epsilon$ provides stronger privacy guarantees but at the expense of increased noise, potentially reducing data utility. A larger $\epsilon$ allows for less noise, thus preserving more of the characteristics of original data but with weaker privacy protection. We calibrate $\epsilon$ to find the optimal balance between privacy and utility, taking into account the context of the data. For instance, with general topics, we might allow for more noise, whereas in highly sensitive contexts, we would choose a higher $\epsilon$ to maintain data fidelity while still offering privacy protection. \\

\section{Experimental Evaluation}

In this section, we present a comprehensive evaluation of our synthetic data generation methodology. We assess whether synthetic datasets produced via our Differential Privacy (DP)-augmented Large Language Model (LLM) approach can serve as effective substitutes for real data in both traditional machine learning (ML) tasks and in-context learning (ICL) scenarios. The evaluation involves training and testing various ML models, as well as conducting ICL experiments with two distinct LLM architectures.
To mitigate the computational and temporal constraints associated with this preliminary investigation, we conducted experiments on a sampled subset of the AGNews dataset rather than employing the full dataset. This choice allows us to efficiently identify key trends and insights regarding the privacy-utility trade-off in synthetic data use while laying the groundwork for future, more extensive studies.

By employing LLM in this nuanced methodology, we aim to generate synthetic datasets that are both functional for further machine learning tasks and compliant with stringent privacy requirements, ensuring that any sensitive information from the original dataset remains confidential. 


We discussed our hyperparameter settings in Apperndix \ref{appendix:implementation}.

\subsection{Dataset}
Our project aims to evaluate the utility of Differential Privacy (DP)-augmented synthetic datasets for machine learning tasks, prioritizing privacy.In our experimental evaluation, we focused on reporting results for the AGNews \cite{zhang2015character} dataset.  

We selected the AGNews dataset as our primary evaluation platform because it is well-studied, contains a variety of news topics, and supports straightforward text classification. In our experimental setting we used 12000 samples for trainings the models and 4000 samples for testing. 


\subsection{Clarifying the ML Task and the Role of the LLM} Our primary ML task is \emph{text classification}, specifically categorizing news articles into four classes (World, Sports, Business, Sci/Tech) using the AGNews dataset \cite{zhang2015character}. In a typical ML setup, a model is trained directly on thousands of labeled examples until it learns to generalize. For instance, a model might see multiple examples of “Business” headlines and learn words and phrases associated with financial reports, corporate earnings, or market indicators.
In contrast, an LLM can perform classification through \emph{prompting} and \emph{in-context learning (ICL)}. Rather than training the model’s parameters extensively, we present the LLM with a few carefully chosen examples at inference time. For example, we show the LLM a prompt:
\begin{verbatim} Classify the headline into 
[World, Sports, Business, Sci/Tech].
Example: "Government imposes new tariffs 
on foreign goods" -> Business 
"Local team clinches regional 
championship" -> Sports
Now classify: "Stock markets rally after 
positive economic indicators" \end{verbatim}
The LLM uses these provided examples to infer the correct category (likely “Business”) without additional training. However, if these demonstration examples contain sensitive data, the model might memorize and reveal private details. Our DP-based synthetic data prevents such leakage by ensuring that no individual piece of sensitive information is directly represented.

\subsection{Model Choices and Experimental Setup} 

To thoroughly assess the utility of the synthetic data , we evaluated a range of models with differing complexities and requirements: 

\begin{itemize} 

    \item \textbf{Multinomial Naive Bayes (MNB):} A simple model that relies on frequency counts of words. If MNB performs reasonably on synthetic data, it suggests that core statistical properties are preserved \cite{10.1007/978-3-540-30549-1_43}. 
    
    \item \textbf{Support Vector Machine (SVM):} SVM can handle high-dimensional feature spaces effectively. Its performance on synthetic data indicates whether essential discriminative features remain intact \cite{708428}. 
    
    \item \textbf{Gated Recurrent Unit (GRU) and Long Short-Term Memory (LSTM):} These recurrent neural networks leverage sequential and contextual information. If these models show reasonable performance, it means the synthetic data retains more complex patterns, not just basic word frequencies \cite{10.1162/neco.1997.9.8.1735, chung2014empirical}. 
    
    \item \textbf{LLM-based ICL (gpt-4o-mini and gemini-1.5-flash):} By testing in-context learning with synthetic data as demonstration examples, we assess whether the synthetic data can guide an LLM to make correct classifications, reflecting higher-level semantic coherence. 

\end{itemize}

Each of these model types addresses a different layer of complexity in text classification. From basic statistical features (MNB, SVM) to advanced temporal patterns (GRU, LSTM) and finally to context-based reasoning (ICL with LLMs), this multi-tiered approach provides a comprehensive understanding of how well our synthetic data stands in for real data. For word embedding, we employ TfidfVectorizer to prepare the data for training Multinomial Naive Bayes (MNB) and Support Vector Machine (SVM) models. This choice is justified by TfidfVectorizer's ability to reflect the importance of words in the context of the entire corpus, which is particularly useful for traditional machine learning models that benefit from frequency-based features. For the GRU and LSTM models, we use GloVe \cite{pennington-etal-2014-glove} embeddings to preprocess the training data. GloVe embeddings capture semantic relationships between words through their co-occurrence statistics, providing a dense vector representation that is well-suited for deep learning models that thrive on understanding contextual and semantic nuances in text. This dual approach allows us to leverage the strengths of each embedding method tailored to the specific requirements of the models we are training.

\subsection{Performance Metrics and Interpretation of Results} 

Our primary evaluation metric is \emph{accuracy}, the percentage of test samples correctly classified. Accuracy is intuitive and sufficient for balanced classification tasks like AGNews. Although accuracy below 90\% may seem disappointing, it is essential to interpret these numbers in the context of privacy goals. Achieving near-random performance (e.g., around 25\% for a four-class problem) would be useless, while reaching moderate accuracy (e.g., above 50\%) already indicates that the synthetic data has captured some meaningful signals. As our method evolves, we aim to improve accuracy while maintaining strong privacy guarantees.

\section{Results}

In this section, we assess the utility of our synthetic data across a range of machine learning models and Large Language Model (LLM)-based approaches.

\subsection{Machine Learning Model Evaluation}
We first assessed how the DP-augmented synthetic data influenced the performance of a range of ML models, including Multinomial Naive Bayes (MNB), Support Vector Machines (SVM), Gated Recurrent Units (GRU), and Long Short-Term Memory (LSTM) networks. In Table~\ref{tab:agenws-ml}, we report the accuracy of each model trained on either the original AGNews dataset or on our DP-based synthetic data. To establish a reference point, we began by training all models on the original dataset, then repeated the training with the synthetic data generated via our proposed LLM based approach.

\begin{table}[h]
\centering
\begin{tabular}{lcc}
\hline
\textbf{Method} & \begin{tabular}{c} \textbf{Accuracy} \\ \textbf{Original Data} \end{tabular} & \begin{tabular}{c} \textbf{Accuracy} \\ \textbf{Synthetic Data} \end{tabular} \\ \hline \hline
MNB & 80.73 \%  & 77.92 \% \\ \hline
SVM  & 86.75 \%  & 76.43 \% \\ \hline
GRU   & 83.62 \% & 65.62 \% \\ \hline 
LSTM & 84.42 \% & 64.83 \%  \\ \hline 

\end{tabular}
\caption{Performance Comparison of Machine Learning Models on Original vs. Synthetic Data. This table evaluates the effectiveness of various machine learning models (including both simple and complex architectures) trained on synthetic data versus the original AGNews dataset. It highlights the utility of synthetic data for machine learning by comparing classification accuracy across different model types.}
\label{tab:agenws-ml}
\end{table}
\begin{table*}[ht]
\centering
\begin{tabular}{lccccc}
\hline
\textbf{Method} & \textbf{Zero-shot} & \multicolumn{2}{c}{\textbf{Accuracy (2-shot)}} & \multicolumn{2}{c}{\textbf{Accuracy (4-shot)}} \\
\cline{3-6}
& & \textbf{Original} & \textbf{Synthetic} & \textbf{Original} & \textbf{Synthetic} \\
\hline \hline
LLM-ICL (gpt-4o-mini) & 58.67\% & 77.28 \% & 72.42 \% & 81.37 \% & 75.43 \% \\ \hline
LLM-ICL (gemini-1.5-flash) & 54.76\% & 65.71 \% & 61.21 \% & 72.83 \% & 69.28 \% \\ \hline
\end{tabular}
\caption{Comparison of LLM In-Context Learning Accuracy with Original vs. Our generated Synthetic Data. This table showcases the accuracy of Large Language Models (LLMs) when performing In-Context Learning (ICL) tasks using both the original AGNews dataset and our privacy-preserving synthetic data. It quantifies how well synthetic data can substitute for real data in terms of model performance for zero-shot, 2-shot, and 4-shot learning scenarios.}
\label{tab:agenws-llm}
\end{table*}

For the simpler ML models, MNB and SVM, the accuracy decline is comparatively modest (around 3\% and 10\%, respectively), indicating that the synthetic data preserves sufficient statistical features to enable effective classification. This suggests that our generation method can produce datasets that remain useful for tasks where interpretability and efficiency are paramount, even if the ultimate performance is somewhat reduced.

In contrast, the sequence-based GRU and LSTM architectures exhibit substantially larger accuracy reductions. These models rely heavily on subtle semantic and temporal patterns, which may be partially obscured by the DP noise and the synthetic generation process. This shortfall highlights a current limitation in our approach for tasks requiring sophisticated contextual understanding. Future refinements in noise calibration and generation strategies may be necessary to better capture the nuanced structures these models depend upon.


\subsection{LLM In-Context Learning Performance}

The Table \ref{tab:agenws-llm} provides a detailed comparison of how Large Language Models (LLMs), specifically gpt-4o-mini and gemini-1.5-flash, perform across zero-shot, 2-shot, and 4-shot in-context learning scenarios, with both original and synthetic data from the AGNews dataset. In zero-shot learning, where no specific examples are provided, the models still manage to achieve accuracies around 58.67\% and 54.76\% respectively, demonstrating a foundational understanding derived from their pre-training. This baseline performance, however, is markedly lower than when contextual examples are provided, underscoring the effectiveness of in-context learning.

As we move from 2-shot to 4-shot learning, there is a notable enhancement in accuracy for both LLMs, on both types of data. This suggests that the more context we provide, the better these models can adapt their responses to mimic the desired task output. However, there is an evident performance gap between models trained on original versus synthetic data, with synthetic data consistently yielding lower accuracies. This gap becomes less significant as we increase the number of shots, indicating that additional context can compensate somewhat for the synthetic data's limitations in capturing the real data's nuances.

Comparing the two models, gpt-4o-mini generally outperforms gemini-1.5-flash, which could be due to differences in their pre-training, model architecture, or fine-tuning. Despite this, both models demonstrate that synthetic data can be effectively utilized in in-context learning, particularly when privacy concerns limit the use of real data. 

The trade-off between privacy and accuracy is clear; synthetic data introduces a performance penalty, but this can be mitigated to an extent by providing more contextual examples. This analysis confirms that while synthetic data might not replicate the performance of original data, it offers significant utility in privacy-sensitive applications. Future research could look into refining synthetic data generation or exploring further optimizations in in-context learning to narrow the performance gap further.

\subsection{Evaluating Utility and Privacy Trade-Off}

 To further investigate the relationship between privacy and utility, we experimented with different privacy budgets ($\epsilon$) controlling the intensity of DP noise. As shown in Table~\ref{tab:agenws-llm-epsilon}, lowering $\epsilon$ (stronger privacy) results in more pronounced accuracy degradation, while increasing $\epsilon$ (weaker privacy) improves performance. This outcome aligns with the fundamental DP trade-off: protecting individuals more thoroughly reduces the clarity of patterns in the data, thereby hampering ML performance.

\begin{table}[h]
\centering
\begin{tabular}{lcc}
\hline
\textbf{Method}    & \textbf{Accuracy} \\  \hline \hline
LLM ICL ($\epsilon = 0$)  & 69.83 \% \\ \hline
LLM ICL ($\epsilon = 0.5$) & 72.43 \% \\ \hline 
LLM ICL ($\epsilon = 1$) & 73.64 \%  \\ \hline 
LLM ICL ($\epsilon = 10$)  & 75.42 \%  \\ \hline 
\end{tabular}
\caption{Impact of Differential Privacy Levels on LLM ICL Accuracy Using Synthetic Data. This table explores how varying levels of Differential Privacy (DP), with various $\epsilon$, affect the accuracy of Large Language Models (LLMs) in in-context learning tasks ( $\epsilon = 0, 0.5, 1,$ and $10$). It demonstrates the trade-off between privacy and utility, showing how accuracy varies as privacy protections are either intensified or reduced. Notably, even with privacy constraints, LLM-based in-context learning with synthetic data achieves performance comparable to that using the original data.}
\label{tab:agenws-llm-epsilon}
\end{table}

Though these accuracies may not match those of models trained directly on real data, they must be considered in the context of privacy preservation. Achieving perfect accuracy is not the sole objective in scenarios where personal information must remain confidential. Instead, identifying a balance where the ML model offers practical utility while meeting strict privacy requirements remains the overarching goal.

\subsection{Privacy Assurance in Synthetic Data Generation}

Our method for generating synthetic data is designed to ensure privacy by fundamentally avoiding the use of original data in the training phase. Instead, we employ a Large Language Model (LLM) to generate new data based on a few in-context examples, which serve only as guidance rather than direct sources of information. This approach means the LLM does not learn from or memorize the sensitive content of the original dataset during the generation process. 

To further safeguard privacy, we incorporate noise into the synthetic data through Differential Privacy mechanisms like Laplace or Gaussian noise. This noise ensures that even if the synthetic data structurally or thematically resembles the original data, it is textually distant enough to prevent any direct linkage back to individual entries in the original dataset. By doing so, we maintain the statistical and semantic properties necessary for machine learning tasks while ensuring that the generated data does not compromise the privacy of the source data. This dual strategy of not training on original data and adding noise effectively protects privacy, making our synthetic dataset a privacy-preserving alternative for data-driven applications.

\subsection{Interpreting the Findings and Future Directions} These results confirm that while DP-augmented synthetic data does not fully replicate the richness of the original dataset, it can still enable non-trivial classification performance. Simpler models and tasks are more forgiving of the noise-induced distortions, whereas models requiring complex semantic or temporal understanding face greater challenges.

For LLM-driven ICL, increasing the number of examples mitigates some of the losses, indicating potential strategies to refine prompting techniques. Moreover, systematically tuning $\epsilon$ values and experimenting with alternative noise distributions (e.g., Laplace) offers pathways to optimize this trade-off. As the field progresses, integrating more sophisticated generative techniques, exploring data augmentation, or leveraging domain-specific priors may help retain more subtle patterns while maintaining robust privacy guarantees.

Our experiments indicate that the DP augmented LLM approach can generate synthetic data that, provides meaningful support for various ML and ICL tasks. Although improvements are needed to better capture complex dependencies and raise accuracy levels, the fundamental promise of privacy-preserving synthetic data as a safer alternative to using real, sensitive datasets is evident. The key is in balancing privacy needs with the required data utility, understanding that some performance degradation is a trade-off for enhanced privacy protection. This balance will inform future synthetic data strategies in machine learning and data science.





\section{Conclusion} 
This work demonstrates that integrating Differential Privacy (DP) directly into Large Language Model (LLM)-based data generation can produce privacy-preserving synthetic datasets that retain sufficient utility for certain ML and in-context learning tasks. By tuning parameters such as the privacy budget $\epsilon$, stakeholders can dynamically balance data utility with strict privacy requirements. Although the current approach does not fully capture the semantic and temporal nuances of original data, our results establish a promising baseline for privacy-preserving synthetic data generation in sensitive domains.
Future directions include refining DP strategies to better preserve complex data patterns, employing more diverse evaluation metrics, and exploring adaptive or data-driven noise mechanisms. Additionally, advancing prompt engineering and conducting targeted evaluations against a broader array of privacy attacks will further solidify the robustness of this method. As privacy regulations evolve, ensuring compliance and maintaining ethical standards remain paramount. By pursuing these enhancements, this research moves toward a framework where data-driven insights can be safely harnessed without compromising individual privacy.

\section*{Limitations}

While our study introduces a novel method for generating privacy-preserving synthetic data, it is crucial to acknowledge several limitations that might impact how our findings are interpreted and applied. 

The quality and utility of the synthetic data generated by LLMs, such as gpt-4o-mini, are contingent upon the performance of model and training. Furthermore, there is the issue of scalability and computational resources. Our approach bypasses the need for traditional model training for each new dataset, yet generating synthetic data still demands significant computational power, particularly with larger datasets or for tasks requiring intricate data simulation. This might restrict the practical implementation of our method in settings with limited resources.

Another limitation is the trade-off between privacy and data utility. We have demonstrated how adjusting the $\epsilon$ parameter can balance these aspects, but determining the optimal privacy level for various applications remains a challenge. Increasing privacy protections often comes at the cost of reduced data utility, which could render synthetic data less effective for applications needing high accuracy or dealing with detailed data nuances.

Our evaluation was confined to the AGNews dataset, which, while suitable for news classification, might not reflect the behavior of our method across all data types or domains. This specificity to one dataset could limit the generalizability of our findings. 

There is also a theoretical risk of information leakage despite employing differential privacy. If the noise added is not appropriately calibrated for the dataset or if new, more sophisticated attack methods emerge, the privacy protections might be undermined.

\bibliography{custom}

\begin{thebibliography}{40}
\providecommand{\natexlab}[1]{#1}

\bibitem[{Abadi et~al.(2016)Abadi, Chu, Goodfellow, McMahan, Mironov, Talwar, and Zhang}]{10.1145/2976749.2978318}
Martin Abadi, Andy Chu, Ian Goodfellow, H.~Brendan McMahan, Ilya Mironov, Kunal Talwar, and Li~Zhang. 2016.
\newblock \href {https://doi.org/10.1145/2976749.2978318} {Deep learning with differential privacy}.
\newblock In \emph{Proceedings of the 2016 ACM SIGSAC Conference on Computer and Communications Security}, CCS '16, page 308–318, New York, NY, USA. Association for Computing Machinery.

\bibitem[{Achiam et~al.(2023)Achiam, Adler, Agarwal, Ahmad, Akkaya, Aleman, Almeida, Altenschmidt, Altman, Anadkat et~al.}]{achiam2023gpt}
Josh Achiam, Steven Adler, Sandhini Agarwal, Lama Ahmad, Ilge Akkaya, Florencia~Leoni Aleman, Diogo Almeida, Janko Altenschmidt, Sam Altman, Shyamal Anadkat, et~al. 2023.
\newblock Gpt-4 technical report.
\newblock \emph{arXiv preprint arXiv:2303.08774}.

\bibitem[{Arnold and Neunhoeffer(2020)}]{arnold2020differentially}
Q.~Arnold and T.~Neunhoeffer. 2020.
\newblock \href {https://arxiv.org/abs/2011.05537} {Differentially private synthetic data: Applied evaluations and enhancements}.
\newblock \emph{arXiv preprint arXiv:2011.05537}.

\bibitem[{Brown et~al.(2020{\natexlab{a}})Brown, Mann, Ryder, Subbiah, Kaplan, Dhariwal, Neelakantan, Shyam, Sastry, Askell, Agarwal, Herbert-Voss, Krueger, Henighan, Child, Ramesh, Ziegler, Wu, Winter, Hesse, Chen, Sigler, Litwin, Gray, Chess, Clark, Berner, McCandlish, Radford, Sutskever, and Amodei}]{NEURIPS2020_1457c0d6}
Tom Brown, Benjamin Mann, Nick Ryder, Melanie Subbiah, Jared~D Kaplan, Prafulla Dhariwal, Arvind Neelakantan, Pranav Shyam, Girish Sastry, Amanda Askell, Sandhini Agarwal, Ariel Herbert-Voss, Gretchen Krueger, Tom Henighan, Rewon Child, Aditya Ramesh, Daniel Ziegler, Jeffrey Wu, Clemens Winter, Chris Hesse, Mark Chen, Eric Sigler, Mateusz Litwin, Scott Gray, Benjamin Chess, Jack Clark, Christopher Berner, Sam McCandlish, Alec Radford, Ilya Sutskever, and Dario Amodei. 2020{\natexlab{a}}.
\newblock \href {https://proceedings.neurips.cc/paper_files/paper/2020/file/1457c0d6bfcb4967418bfb8ac142f64a-Paper.pdf} {Language models are few-shot learners}.
\newblock In \emph{Advances in Neural Information Processing Systems}, volume~33, pages 1877--1901. Curran Associates, Inc.

\bibitem[{Brown et~al.(2020{\natexlab{b}})Brown, Mann, Ryder, Subbiah, Kaplan, Dhariwal, Neelakantan, Shyam, Sastry, Askell, Agarwal, Herbert-Voss, Krueger, Henighan, Child, Ramesh, Ziegler, Wu, Winter, Hesse, Chen, Sigler, Litwin, Gray, Chess, Clark, Berner, McCandlish, Radford, Sutskever, and Amodei}]{brown2020languagemodelsfewshotlearners}
Tom~B. Brown, Benjamin Mann, Nick Ryder, Melanie Subbiah, Jared Kaplan, Prafulla Dhariwal, Arvind Neelakantan, Pranav Shyam, Girish Sastry, Amanda Askell, Sandhini Agarwal, Ariel Herbert-Voss, Gretchen Krueger, Tom Henighan, Rewon Child, Aditya Ramesh, Daniel~M. Ziegler, Jeffrey Wu, Clemens Winter, Christopher Hesse, Mark Chen, Eric Sigler, Mateusz Litwin, Scott Gray, Benjamin Chess, Jack Clark, Christopher Berner, Sam McCandlish, Alec Radford, Ilya Sutskever, and Dario Amodei. 2020{\natexlab{b}}.
\newblock \href {https://arxiv.org/abs/2005.14165} {Language models are few-shot learners}.
\newblock \emph{Preprint}, arXiv:2005.14165.

\bibitem[{Cai et~al.(2021)Cai, Lei, Wei, and Xiao}]{10.14778/3476249.3476272}
Kuntai Cai, Xiaoyu Lei, Jianxin Wei, and Xiaokui Xiao. 2021.
\newblock \href {https://doi.org/10.14778/3476249.3476272} {Data synthesis via differentially private markov random fields}.
\newblock \emph{Proc. VLDB Endow.}, 14(11):2190–2202.

\bibitem[{Chua et~al.(2024)Chua, Ghazi, Huang, Kamath, Kumar, Liu, Manurangsi, Sinha, and Zhang}]{chua2024mind}
Lynn Chua, Badih Ghazi, Yangsibo Huang, Pritish Kamath, Ravi Kumar, Daogao Liu, Pasin Manurangsi, Amer Sinha, and Chiyuan Zhang. 2024.
\newblock \href {https://openreview.net/forum?id=Jd0bCD12DS} {Mind the privacy unit! user-level differential privacy for language model fine-tuning}.
\newblock In \emph{First Conference on Language Modeling}.

\bibitem[{Chung et~al.(2014)Chung, Gulcehre, Cho, and Bengio}]{chung2014empirical}
Junyoung Chung, Caglar Gulcehre, KyungHyun Cho, and Yoshua Bengio. 2014.
\newblock Empirical evaluation of gated recurrent neural networks on sequence modeling.
\newblock \emph{arXiv preprint arXiv:1412.3555}.

\bibitem[{Dwork et~al.(2014)Dwork, Roth et~al.}]{dwork2014algorithmic}
Cynthia Dwork, Aaron Roth, et~al. 2014.
\newblock The algorithmic foundations of differential privacy.
\newblock \emph{Foundations and Trends{\textregistered} in Theoretical Computer Science}, 9(3--4):211--407.

\bibitem[{Goodfellow et~al.(2014)Goodfellow, Pouget-Abadie, Mirza, Xu, Warde-Farley, Ozair, Courville, and Bengio}]{goodfellow2014generative}
Ian Goodfellow, Jean Pouget-Abadie, Mehdi Mirza, Bing Xu, David Warde-Farley, Sherjil Ozair, Aaron Courville, and Yoshua Bengio. 2014.
\newblock Generative adversarial nets.
\newblock \emph{Advances in neural information processing systems}, 27.

\bibitem[{Hardt et~al.(2012)Hardt, Ligett, and Mcsherry}]{NIPS2012_208e43f0}
Moritz Hardt, Katrina Ligett, and Frank Mcsherry. 2012.
\newblock \href {https://proceedings.neurips.cc/paper_files/paper/2012/file/208e43f0e45c4c78cafadb83d2888cb6-Paper.pdf} {A simple and practical algorithm for differentially private data release}.
\newblock In \emph{Advances in Neural Information Processing Systems}, volume~25. Curran Associates, Inc.

\bibitem[{Hearst et~al.(1998)Hearst, Dumais, Osuna, Platt, and Scholkopf}]{708428}
M.A. Hearst, S.T. Dumais, E.~Osuna, J.~Platt, and B.~Scholkopf. 1998.
\newblock \href {https://doi.org/10.1109/5254.708428} {Support vector machines}.
\newblock \emph{IEEE Intelligent Systems and their Applications}, 13(4):18--28.

\bibitem[{Hochreiter and Schmidhuber(1997)}]{10.1162/neco.1997.9.8.1735}
Sepp Hochreiter and J\"{u}rgen Schmidhuber. 1997.
\newblock \href {https://doi.org/10.1162/neco.1997.9.8.1735} {Long short-term memory}.
\newblock \emph{Neural Comput.}, 9(8):1735–1780.

\bibitem[{Hong et~al.(2024)Hong, Wang, Zhang, LI, Li, and Wang}]{hong2024dpopt}
Junyuan Hong, Jiachen~T. Wang, Chenhui Zhang, Zhangheng LI, Bo~Li, and Zhangyang Wang. 2024.
\newblock \href {https://openreview.net/forum?id=Ifz3IgsEPX} {{DP}-{OPT}: Make large language model your privacy-preserving prompt engineer}.
\newblock In \emph{The Twelfth International Conference on Learning Representations}.

\bibitem[{Hurst et~al.(2024)Hurst, Lerer, Goucher, Perelman, Ramesh, Clark, Ostrow, Welihinda, Hayes, Radford et~al.}]{hurst2024gpt}
Aaron Hurst, Adam Lerer, Adam~P Goucher, Adam Perelman, Aditya Ramesh, Aidan Clark, AJ~Ostrow, Akila Welihinda, Alan Hayes, Alec Radford, et~al. 2024.
\newblock Gpt-4o system card.
\newblock \emph{arXiv preprint arXiv:2410.21276}.

\bibitem[{Keshta and Odeh(2021)}]{keshta2021security}
Ismail Keshta and Ammar Odeh. 2021.
\newblock Security and privacy of electronic health records: Concerns and challenges.
\newblock \emph{Egyptian Informatics Journal}, 22(2):177--183.

\bibitem[{Kibriya et~al.(2004)Kibriya, Frank, Pfahringer, and Holmes}]{10.1007/978-3-540-30549-1_43}
Ashraf~M. Kibriya, Eibe Frank, Bernhard Pfahringer, and Geoffrey Holmes. 2004.
\newblock \href {https://doi.org/10.1007/978-3-540-30549-1_43} {Multinomial naive bayes for text categorization revisited}.
\newblock In \emph{Proceedings of the 17th Australian Joint Conference on Advances in Artificial Intelligence}, AI'04, page 488–499, Berlin, Heidelberg. Springer-Verlag.

\bibitem[{Kingma et~al.(2019)Kingma, Welling et~al.}]{kingma2019introduction}
Diederik~P Kingma, Max Welling, et~al. 2019.
\newblock An introduction to variational autoencoders.
\newblock \emph{Foundations and Trends{\textregistered} in Machine Learning}, 12(4):307--392.

\bibitem[{Liu et~al.(2024)Liu, Wei, Liu, Si, Zhang, Rao, Zheng, Peng, Yang, Zhou et~al.}]{liu2024best}
Ruibo Liu, Jerry Wei, Fangyu Liu, Chenglei Si, Yanzhe Zhang, Jinmeng Rao, Steven Zheng, Daiyi Peng, Diyi Yang, Denny Zhou, et~al. 2024.
\newblock \href {https://arxiv.org/abs/2404.07503} {Best practices and lessons learned on synthetic data for language models}.
\newblock \emph{arXiv e-prints}, pages arXiv--2404.

\bibitem[{Long et~al.(2021)}]{long2021g_pate}
Y.~Long et~al. 2021.
\newblock \href {https://papers.nips.cc/paper/2021/file/02d6c61c1442118e8b0f97e2c90f76d7-Paper.pdf} {G-pate: Scalable differentially private data generator via private aggregation of teacher discriminators}.
\newblock In \emph{Proceedings of the 35th Conference on Neural Information Processing Systems (NeurIPS 2021)}, pages 2965--2977. Neural Information Processing Systems Foundation.

\bibitem[{Marjanov et~al.(2023)Marjanov, Konstantinou, J{\'o}{\'z}wiak, and Spagnuelo}]{marjanov2023data}
Tina Marjanov, Maria Konstantinou, Magdalena J{\'o}{\'z}wiak, and Dayana Spagnuelo. 2023.
\newblock Data security on the ground: Investigating technical and legal requirements under the gdpr.
\newblock \emph{Proceedings on Privacy Enhancing Technologies}.

\bibitem[{Nahid(2024)}]{nahid2024improving}
Md~Mahadi~Hasan Nahid. 2024.
\newblock \href {https://doi.org/10.7939/r3-ckmh-a783} {Improving table reasoning through table decomposition and normalization}.
\newblock \emph{Univerity of Alberta}.

\bibitem[{Nahid and Rafiei(2024{\natexlab{a}})}]{nahid-rafiei-2024-normtab}
Md~Mahadi~Hasan Nahid and Davood Rafiei. 2024{\natexlab{a}}.
\newblock \href {https://doi.org/10.18653/v1/2024.findings-emnlp.203} {{N}orm{T}ab: Improving symbolic reasoning in {LLM}s through tabular data normalization}.
\newblock In \emph{Findings of the Association for Computational Linguistics: EMNLP 2024}, pages 3569--3585, Miami, Florida, USA. Association for Computational Linguistics.

\bibitem[{Nahid and Rafiei(2024{\natexlab{b}})}]{nahid-rafiei-2024-tabsqlify}
Md~Mahadi~Hasan Nahid and Davood Rafiei. 2024{\natexlab{b}}.
\newblock \href {https://doi.org/10.18653/v1/2024.naacl-long.320} {{T}ab{SQL}ify: Enhancing reasoning capabilities of {LLM}s through table decomposition}.
\newblock In \emph{Proceedings of the 2024 Conference of the North American Chapter of the Association for Computational Linguistics: Human Language Technologies (Volume 1: Long Papers)}, pages 5725--5737, Mexico City, Mexico. Association for Computational Linguistics.

\bibitem[{Nayeem and Rafiei(2024)}]{nayeem-rafiei-2024-kidlm}
Mir~Tafseer Nayeem and Davood Rafiei. 2024.
\newblock \href {https://doi.org/10.18653/v1/2024.emnlp-main.277} {{K}id{LM}: Advancing language models for children {--} early insights and future directions}.
\newblock In \emph{Proceedings of the 2024 Conference on Empirical Methods in Natural Language Processing}, pages 4813--4836, Miami, Florida, USA. Association for Computational Linguistics.

\bibitem[{Nouwens et~al.(2020)Nouwens, Liccardi, Veale, Karger, and Kagal}]{nouwens2020dark}
Midas Nouwens, Ilaria Liccardi, Michael Veale, David Karger, and Lalana Kagal. 2020.
\newblock Dark patterns after the gdpr: Scraping consent pop-ups and demonstrating their influence.
\newblock In \emph{Proceedings of the 2020 CHI conference on human factors in computing systems}, pages 1--13.

\bibitem[{Pennington et~al.(2014)Pennington, Socher, and Manning}]{pennington-etal-2014-glove}
Jeffrey Pennington, Richard Socher, and Christopher Manning. 2014.
\newblock \href {https://doi.org/10.3115/v1/D14-1162} {{G}lo{V}e: Global vectors for word representation}.
\newblock In \emph{Proceedings of the 2014 Conference on Empirical Methods in Natural Language Processing ({EMNLP})}, pages 1532--1543, Doha, Qatar. Association for Computational Linguistics.

\bibitem[{Rosenblatt et~al.(2023)Rosenblatt, Herman, Holovenko, Lee, Loftus, McKinnie, Rumezhak, Stadnik, Howe, and Stoyanovich}]{10.14778/3611479.3611517}
Lucas Rosenblatt, Bernease Herman, Anastasia Holovenko, Wonkwon Lee, Joshua Loftus, Elizabeth McKinnie, Taras Rumezhak, Andrii Stadnik, Bill Howe, and Julia Stoyanovich. 2023.
\newblock \href {https://doi.org/10.14778/3611479.3611517} {Epistemic parity: Reproducibility as an evaluation metric for differential privacy}.
\newblock \emph{Proc. VLDB Endow.}, 16(11):3178–3191.

\bibitem[{Rosenblatt et~al.(2020)Rosenblatt, Liu, Pouyanfar, de~Leon, Desai, and Allen}]{rosenblatt2020differentiallyprivatesyntheticdata}
Lucas Rosenblatt, Xiaoyan Liu, Samira Pouyanfar, Eduardo de~Leon, Anuj Desai, and Joshua Allen. 2020.
\newblock \href {https://arxiv.org/abs/2011.05537} {Differentially private synthetic data: Applied evaluations and enhancements}.
\newblock \emph{Preprint}, arXiv:2011.05537.

\bibitem[{Samarin et~al.(2023)Samarin, Kothari, Siyed, Bjorkman, Yuan, Wijesekera, Alomar, Fischer, Hoofnagle, and Egelman}]{samarin2023lessons}
Nikita Samarin, Shayna Kothari, Zaina Siyed, Oscar Bjorkman, Reena Yuan, Primal Wijesekera, Noura Alomar, Jordan Fischer, Chris Hoofnagle, and Serge Egelman. 2023.
\newblock Lessons in vcr repair: Compliance of android app developers with the california consumer privacy act (ccpa).
\newblock \emph{Proceedings on Privacy Enhancing Technologies}, 3:103--121.

\bibitem[{Shokri et~al.(2017)Shokri, Stronati, Song, and Shmatikov}]{shokri2017membership}
Reza Shokri, Marco Stronati, Congzheng Song, and Vitaly Shmatikov. 2017.
\newblock Membership inference attacks against machine learning models.
\newblock In \emph{2017 IEEE symposium on security and privacy (SP)}, pages 3--18. IEEE.

\bibitem[{Stadler et~al.(2022)Stadler, Oprisanu, and Troncoso}]{stadler2022synthetic}
Theresa Stadler, Bristena Oprisanu, and Carmela Troncoso. 2022.
\newblock \href {https://www.usenix.org/conference/usenixsecurity22/presentation/stadler} {Synthetic data – anonymisation groundhog day}.
\newblock In \emph{31st USENIX Security Symposium (USENIX Security 22)}, pages 1451--1468, Boston, MA. USENIX Association.

\bibitem[{Tang et~al.(2024)Tang, Shin, Inan, Manoel, Mireshghallah, Lin, Gopi, Kulkarni, and Sim}]{tang2024privacypreserving}
Xinyu Tang, Richard Shin, Huseyin~A Inan, Andre Manoel, Fatemehsadat Mireshghallah, Zinan Lin, Sivakanth Gopi, Janardhan Kulkarni, and Robert Sim. 2024.
\newblock \href {https://openreview.net/forum?id=oZtt0pRnOl} {Privacy-preserving in-context learning with differentially private few-shot generation}.
\newblock In \emph{The Twelfth International Conference on Learning Representations}.

\bibitem[{Team et~al.(2024)Team, Georgiev, Lei, Burnell, Bai, Gulati, Tanzer, Vincent, Pan, Wang et~al.}]{team2024gemini}
Gemini Team, Petko Georgiev, Ving~Ian Lei, Ryan Burnell, Libin Bai, Anmol Gulati, Garrett Tanzer, Damien Vincent, Zhufeng Pan, Shibo Wang, et~al. 2024.
\newblock Gemini 1.5: Unlocking multimodal understanding across millions of tokens of context.
\newblock \emph{arXiv preprint arXiv:2403.05530}.

\bibitem[{Torkzadehmahani et~al.(2019)Torkzadehmahani, Kairouz, and Paten}]{torkzadehmahani2019dp}
Reihaneh Torkzadehmahani, Peter Kairouz, and Benedict Paten. 2019.
\newblock Dp-cgan: Differentially private synthetic data and label generation.
\newblock In \emph{2019 IEEE/CVF Conference on Computer Vision and Pattern Recognition Workshops (CVPRW)}, pages 98--104. IEEE Computer Society.

\bibitem[{Vietri et~al.(2020)Vietri, Tian, Bun, Steinke, and Wu}]{pmlr-v119-vietri20b}
Giuseppe Vietri, Grace Tian, Mark Bun, Thomas Steinke, and Steven Wu. 2020.
\newblock \href {https://proceedings.mlr.press/v119/vietri20b.html} {New oracle-efficient algorithms for private synthetic data release}.
\newblock In \emph{Proceedings of the 37th International Conference on Machine Learning}, volume 119 of \emph{Proceedings of Machine Learning Research}, pages 9765--9774. PMLR.

\bibitem[{Wei et~al.(2022)Wei, Wang, Schuurmans, Bosma, Xia, Chi, Le, Zhou et~al.}]{wei2022chain}
Jason Wei, Xuezhi Wang, Dale Schuurmans, Maarten Bosma, Fei Xia, Ed~Chi, Quoc~V Le, Denny Zhou, et~al. 2022.
\newblock Chain-of-thought prompting elicits reasoning in large language models.
\newblock \emph{Advances in neural information processing systems}, 35:24824--24837.

\bibitem[{Zhang et~al.(2017)Zhang, Cormode, Procopiuc, Srivastava, and Xiao}]{zhang2017privbayes}
J.~Zhang, G.~Cormode, C.M. Procopiuc, D.~Srivastava, and X.~Xiao. 2017.
\newblock \href {https://doi.org/10.1145/3035918.3035919} {{PrivBayes}: Private data release via bayesian networks}.
\newblock In \emph{Proceedings of the 2017 ACM SIGMOD International Conference on Management of Data}, pages 1423--1434, New York, NY. ACM.

\bibitem[{Zhang et~al.(2015)Zhang, Zhao, and LeCun}]{zhang2015character}
Xiang Zhang, Junbo Zhao, and Yann LeCun. 2015.
\newblock Character-level convolutional networks for text classification.
\newblock \emph{Advances in neural information processing systems}, 28.

\bibitem[{Zhang et~al.(2021)Zhang, Wang, Li, Honorio, Backes, He, Chen, and Zhang}]{zhang2021privsyn}
Zhikun Zhang, Tianhao Wang, Ninghui Li, Jean Honorio, Michael Backes, Shibo He, Jiming Chen, and Yang Zhang. 2021.
\newblock \href {https://www.usenix.org/conference/usenixsecurity21/presentation/zhang-zhikun} {{PrivSyn}: Differentially private data synthesis}.
\newblock In \emph{30th USENIX Security Symposium (USENIX Security 21)}, pages 929--946, Boston, MA. USENIX Association.

\end{thebibliography}

\newpage
\appendix

\section{Background}
\label{appendix: background}
This section provides an overview of key concepts necessary for understanding our approach. We begin by explaining how Large Language Models (LLMs) adapt to various tasks through prompting and in-context learning. We then define Differential Privacy (DP) as a framework for formalizing privacy guarantees, and finally discuss synthetic data generation as a method to protect sensitive information while preserving data utility for machine learning (ML) tasks.
\subsection{Prompting and In-Context Learning} Large Language Models (LLMs) are advanced neural network-based models trained on vast amounts of text data. Traditional ML models typically require explicit retraining or fine-tuning when adapting to new tasks, such as text classification (i.e., assigning a category like \emph{Sports} or \emph{Business} to a news article). In contrast, LLMs can often perform new tasks without extensive parameter updates, using techniques known as \emph{prompting} and \emph{in-context learning (ICL)}.
\emph{Prompting} involves providing an LLM with carefully crafted instructions or examples at inference time, guiding it to produce outputs aligned with a given task \cite{brown2020languagemodelsfewshotlearners}. For instance, if the task is text classification, rather than retraining the entire model, we can supply a prompt that outlines the categories and shows a few labeled examples. The LLM then generates answers consistent with these categories.
\emph{In-context learning} (ICL) further reduces training overhead by embedding a small set of demonstrations—usually just a few labeled examples—directly into the query fed to the model \cite{wei2022chain}. The LLM uses these demonstrations as context to infer the desired output format and reasoning steps for the task at hand. In practical terms, if we want the model to classify a piece of text, we insert a few examples of text-label pairs into the prompt. The model observes these examples and tries to classify the new, unlabeled text following the same pattern, often with surprisingly high accuracy given no explicit retraining.
While prompting and ICL can save computational resources and simplify workflows, they also introduce new privacy challenges. If the prompt or the demonstrations contain sensitive information—such as personal identifiers or medical details—the LLM might memorize and reveal them later. This necessitates effective privacy safeguards that prevent inadvertent data leakage.
\subsection{Differential Privacy (DP)} Differential Privacy (DP) provides a mathematically rigorous framework for preserving individual privacy when analyzing or generating data \cite{dwork2014algorithmic}. In simple terms, DP ensures that the presence or absence of any single data record in a dataset only slightly affects the outcome of a computation. By controlling the degree to which any individual record influences the final output, DP protects against adversaries attempting to infer whether a specific person’s data was included.
The key mechanism that enables DP is the deliberate addition of controlled noise to computations. For example, when querying a dataset to compute statistical summaries or when training a model on sensitive information, small amounts of random noise can be injected into intermediate steps. This ensures that outputs do not depend too heavily on any single record, making it difficult to “reverse-engineer” the original data.
A crucial parameter in DP is the \emph{privacy budget}, often denoted by $\epsilon$. A smaller $\epsilon$ implies stronger privacy guarantees but typically introduces more noise and thus may reduce the utility (i.e., accuracy or performance) of the resulting model or dataset. Practitioners must choose $\epsilon$ values that strike a suitable balance between privacy protection and task effectiveness.
\subsection{Synthetic Data Generation} Many ML tasks, including text classification, rely on large, representative datasets. However, using real-world data may pose significant privacy, ethical, or legal challenges—particularly when the data includes sensitive personal information. \emph{Synthetic data generation} addresses this issue by producing artificial datasets that resemble real data in their statistical properties but do not contain actual individual-level records \cite{tang2024privacypreserving}.
Traditional synthetic data generation methods include probabilistic modeling or employing complex generative models, such as Generative Adversarial Networks (GANs) and Variational Autoencoders (VAEs) \cite{goodfellow2014generative, kingma2019introduction}. More recently, LLMs have emerged as powerful tools for producing synthetic text data that captures linguistic patterns and topic distributions \cite{hong2024dpopt, tang2024privacypreserving}. By guiding an LLM with representative examples, it is possible to generate new text instances that appear similar to the original data without reproducing any specific confidential entries.
When combined with Differential Privacy, synthetic data generation can offer both strong privacy guarantees and practical utility. Specifically, one can integrate DP noise mechanisms directly into the synthetic data creation process, ensuring that each generated data point is influenced by many original records, but never strongly enough to reveal an individual’s sensitive information. The resulting DP-augmented synthetic datasets can then be used for ML tasks—such as text classification—without exposing the original private data. This approach is particularly valuable in fields like healthcare or finance, where regulations and ethical considerations prohibit the use of raw sensitive data for model training or evaluation.
\subsection{Linking Concepts to Our Approach} In our research, we combine these three concepts—prompting and in-context learning with LLMs, Differential Privacy, and synthetic data generation—to tackle the challenge of producing privacy-preserving datasets for ML tasks. By understanding how LLMs can be guided through prompting and ICL, we reduce computational overhead. By applying DP, we ensure that no individual’s data is compromised. By generating synthetic data that mimics real samples without revealing them, we maintain data utility while adhering to strict privacy standards.
This integrated approach aims to produce synthetic datasets that are both useful and compliant with rigorous privacy requirements, ultimately supporting safer, more responsible data-driven innovation in sensitive domains.

\section{Implementation}
\label{appendix:implementation}


\subsection{Hyperparameters Settings for LLMs}
We configured the hyperparameters for in-context learning as per the specifications provided in Table \ref{tab:hyper-parameters1}.

\begin{table}[h]
\centering
\begin{tabular}{lcc}
\hline
\textbf{Parameter}   & \textbf{Synthetic Data Generation} \\ \hline \hline
temperature           & 0.7    \\ \hline
top\_p                & 1       \\ \hline
sample\_n             & 1   \\ \hline 
max\_tokens           & 200   \\ \hline 
num\_shots            & 4     \\ \hline

\end{tabular}
\caption{The hyper-parameters we set in Synthetic Data Generation}
\label{tab:hyper-parameters1}
\end{table}

The average text length in the dataset is approximately 200 words. We configured the maximum tokens to 200. To promote diversity in token generation by the LLM, we set a high temperature of 0.7. 

\subsection{ML Model configuration}

In our study, we utilized an SVM classifier with a range of parameters for optimization, defined as:

\begin{verbatim}
svm_parameters = {
    'clf__C': [0.1, 1, 10],
    'clf__gamma': [1, 0.1, 0.01],
    'clf__kernel': ['rbf', 'linear']
}
\end{verbatim}

This configuration allows us to assess performance on both non-linear and linear data distributions by adjusting the regularization parameter \texttt{C}, the kernel 
coefficient \texttt{gamma}, and the kernel type. For the Multinomial Naive Bayes (MNB) classifier, we employed a pipeline defined as:

\begin{verbatim}
mnb_parameters = Pipeline([
    ('tfidf', TfidfVectorizer()),
    ('clf', MultinomialNB())
])
\end{verbatim}

This pipeline incorporates TF-IDF vectorization for feature extraction followed by the MNB classifier. For our dataset split, we allocated 70\% for training and 30\% for validation to compare the performance of both classifiers.
\\ 

\begin{table}[h]
\centering
\begin{tabular}{lcc}
\hline
\textbf{Layer}   & \textbf{Output Shape}  & \textbf{Param \#}\\ \hline \hline
Embedding           & (None, 975, 100) &  9,308,600    \\ \hline
GRU or LSTM       & (None, 32)   & 12,864    \\ \hline
Dense         & (None, 4)   & 132    \\ \hline
\end{tabular}
\caption{Parameters set for GRU and LSTM classifiers}
\label{tab:gru-lstm-param}
\end{table}

For training the GRU and LSTM classifiers, we utilized the settings summarized in Table \ref{tab:gru-lstm-param}. 

For the embedding layer, we utilized GloVe embeddings to convert input text into numerical vectors, which are then fed into the GRU or LSTM layer. The embedding layer outputs sequences of length 975 with each token represented by a 100-dimensional vector, capturing semantic relationships. The subsequent GRU or LSTM layer reduces the dimensionality to 32 units, allowing the model to capture temporal dependencies. We also incorporated a dropout rate of 0.2 to prevent overfitting by randomly setting 20\% of input units to 0 during training. Finally, a dense layer with 4 units corresponds to our four-class classification task, providing the final output predictions. Additionally, we implemented early stopping to halt training when the validation loss ceased to decrease, thus avoiding unnecessary epochs and potential overfitting.

\subsection{Sample Prompt for Synthetic Data Generation}
For the generation of synthetic data for the news classification task, we used the following prompt to guide the Large Language Model (LLM):

\begin{verbatim}
Your task is to generate synthetic 
dataset for news classification task. 
Here is some example:

Title: Wall St. Bears Claw Back Into 
the Black (Reuters) 
Description: Reuters - Short-sellers, 
Wall Street's dwindling\band of 
ultra-cynics, are seeing green again. 
Class Label: "Sci/Tech"

Title: Singh Leads, but Leonard Is 
Following 
Description: Avoiding the late trouble 
that knocked other contenders off 
track, Vijay Singh held a one-stroke 
lead over Justin Leonard heading 
into the final round of the P.G.A. 
Championship.
Class Label: "Sports"

Title: Two visions of Iraq struggle 
to take hold
Description: Fighting in Najaf threatened 
to undermine a conference to choose a 
national assembly.
Class Label: "World"

Title: Dollar Falls to Fresh 
Low Vs Euro (Reuters)
Description: Reuters - The dollar fell 
to a fresh four-week low\versus the 
euro on Monday after a widening of 
the U.S. trade\gap to record levels 
raised worries about capital inflows 
in\the United States and a possible 
slowdown in the economy.
Class Label: "Bussiness"

You need to generate synthetic dataset 
for these classes: "World", "Bussiness", 
"Sports" and "Sci/Tech".

Now generate ###<NUMBER>### different 
synthetic data without any explanation. 
Output should be in json format containing 
"Title", "Description", "Class_Label".
\end{verbatim}

\subsection{In-Context Learning Prompt for LLM Classifier}
The following prompt was used for in-context learning with the Large Language Model (LLM) classifier to predict news class:

\begin{verbatim}
You are a helpful assistant. Your task is to 
predict news class for a given news.
The classes are: 
classes: "World", "Bussiness", "Sports" 
and "Sci/Tech".

Here are some demonstrations:

Title: Breakthrough in Renewable Energy 
Technology
Description: Innovative new technology in 
renewable energy could lead to more efficient 
solar panels and wind turbines.
Class Label: "Sci/Tech"

Title: College Basketball Tournament 
Kicks Off
Description: The much-anticipated college 
basketball tournament has begun, with teams 
vying for the championship title.
Class Label: "Sports"

Title: Cultural Heritage Sites 
Under Threat
Description: Several cultural heritage 
sites around the world are facing 
threats due to climate change 
and urban development.
Class Label: "World"

Title: Tech Stocks Rally After 
Positive Earnings
Description: Tech stocks saw a significant 
rally today following a series of positive 
earnings reports 
from major companies.
Class Label: "Bussiness"

Now predict only the class label 
for the follwoing news:
###<NEW SAMPLE>###

\end{verbatim}



\end{document}